\documentclass[authorproof,IJAT,researchpaper]{fujipressarticle}
\usepackage{graphicx}
\usepackage{cite}
\usepackage{amsmath}
\usepackage{bm}
\usepackage{subcaption}
\usepackage[table]{xcolor}
\usepackage{url}
\usepackage{etoolbox}

\SetVolume{20}
\SetIssueNo{5}
\SetPubYear{2026}

\makeatletter
\def\@textbottom{\vskip \z@ \@plus 1fil}
\newcommand\arxivnote[1]{{\long\def\@makefntext##1{\noindent##1}\footnotetext{#1}}}
\makeatother

\begin{document}

\pagestyle{fujipressstyle}

\title{Moving Object Detection from Moving Camera Using Focus of Expansion Likelihood and Segmentation}
\author{Masahiro Ogawa, Qi An, and Atsushi Yamashita}
\address{Department of Precision Engineering, Graduate School of Engineering, The University of Tokyo\\
Department of Human and Engineered Environmental Studies, Graduate School of Frontier Sciences, The University of Tokyo\\
E-mail: ogawa@robot.t.u-tokyo.ac.jp, anqi@robot.t.u-tokyo.ac.jp, yamashita@robot.t.u-tokyo.ac.jp}
\markboth{M. Ogawa, Q. An, and A. Yamashita}{Moving Object Detection from Moving Camera Using FoELS}
\dates{2026/04/05}{2026/06/29}
\maketitle
\arxivnote{This is the author's accepted manuscript of the article published in the International Journal of Automation Technology, Vol.~20, No.~5, pp.~491--502, 2026, DOI: 10.20965/ijat.2026.p0491. \copyright~Fuji Technology Press Ltd. Distributed under the Creative Commons Attribution-NoDerivatives 4.0 International License (CC BY-ND 4.0).}

\begin{abstract}
    \noindent Separating moving and static objects from a moving camera viewpoint is essential for 3D reconstruction, autonomous navigation, and scene understanding in robotics.
    Existing approaches often rely primarily on optical flow, which struggle to detect moving objects in complex, structured scenes involving camera motion.
    To address this limitation, we propose Focus of Expansion Likelihood and Segmentation (FoELS), a method based on the core idea of integrating both optical flow and texture information.
    FoELS computes the focus of expansion (FoE) from optical flow and derives an initial motion likelihood from the outliers of the FoE computation.
    This likelihood is then fused with a segmentation-based prior to estimate the final moving probability.
    The method effectively handles challenges including complex structured scenes, rotational camera motion, and parallel motion.
    Comprehensive evaluations on the DAVIS 2016 and FBMS-59 datasets, along with real-world traffic videos including parallel, cross-direction, opposite-direction, and crowded scenes, demonstrate its effectiveness and state-of-the-art performance.
\end{abstract}


\section{Introduction}
\label{section:introduction}
\begin{figure}[t]
    \centering
    \includegraphics[width=0.9\columnwidth]{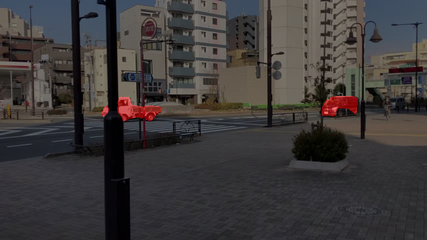}
    \caption{Sample result of FoELS. It detects moving objects from a moving camera at various distances within the scene.}
    \label{fig:result_sample}
\end{figure}

Separating moving objects from static scenes in video is a fundamental task with applications in 3D reconstruction, obstacle avoidance for autonomous vehicles, and scene understanding for assistant robots.
Previous methods \cite{zhangw2020,zhenchenghu1999,yanhchaoyang2019} rely only on optical flow information to differentiate object motion from camera motion.
However, they often fail in complex, structured scenes, with intricate camera motion, or in low-textured environments.
Because flow length depends on an object's relative motion magnitude and distance from the camera,
relying solely on flow makes it difficult to detect moving objects in complex 3D scenes.
This work proposes a novel approach leveraging optical flow and segmentation to overcome these challenges.
As shown in Fig.~\ref{fig:result_sample}, the proposed method, Focus of Expansion Likelihood and Segmentation (FoELS), effectively detects moving objects in complex structured scenes.
Detecting moving objects in dynamic scenes is vital for various robotics applications, such as autonomous navigation and environmental understanding.
While static scene segmentation has advanced significantly, identifying dynamic components remains challenging,
particularly under complex conditions such as rotational motion, camera zoom, and cluttered backgrounds.
The ability to accurately detect moving objects in dynamic scenarios facilitates precise reconstruction of the environment, which is invaluable for augmented and virtual reality applications.

\subsection{Key Challenges}
To detect moving objects from a moving camera, it is necessary to extract motion in the image, and then remove camera-induced motion.
Key challenges in this domain arising from the inherent complexity of real-world environments include:
\begin{enumerate}
    \item Misinterpretation of large optical flow from nearby static objects: Large optical flow magnitudes from close, static objects can be erroneously interpreted as object motion. \label{challenge1}
    \item Insufficient flow in low-textured regions: Environments with minimal texture hinder optical flow algorithms, leading to unreliable motion estimates. \label{challenge2}
    \item Ambiguity in parallel motion: Objects moving parallel to the camera's trajectory often produce optical flow that aligns with the background flow, causing detection ambiguities. \label{challenge3}
    \item Detection of partially stationary objects: Objects with both moving and static parts (e.g., a walking animal with stationary limbs at certain moments) are challenging to classify accurately as moving, yet such distinction is crucial for applications like 3D reconstruction. \label{challenge4}
\end{enumerate}

\subsection{Contributions and Key Ideas to Overcome Challenges}
\label{subsec:contributions}
To address these four challenges, FoELS integrates the following key ideas, representing three distinct contributions:

\textbf{Contribution 1: Introduction of macroscopic perspective.}
Previous approaches all focus on microscopic information (optical flow).
We incorporate an FoE-based approach for microscopic pixel-level analysis, which addresses both straight-ahead camera motion and the misinterpretation of large optical flow from nearby static objects (Challenge \ref{challenge1}).
However, unlike \cite{zhenchenghu2000}, we do not assume a fixed FoE and instead allow it to vary with each frame, enabling us to handle various camera movements.
Critically, we introduce macroscopic information (texture information through segmentation) to the moving object detection problem, addressing the challenges of misinterpreting large optical flow from nearby static objects (Challenge \ref{challenge1}) and insufficient flow in low-textured regions (Challenge \ref{challenge2}), which \cite{yanhchaoyang2019} struggled with.
Additionally, we introduce object-level refinement to extract complete moving objects even when only parts exhibit motion (Challenge \ref{challenge4}), explained in \ref{subsec:object_level_refinement}.
This leverages panoptic segmentation to ensure detection of complete objects rather than individual moving pixels.
This is the first work to introduce segmentation as a macroscopic prior specifically for moving object detection from a moving camera.

\textbf{Contribution 2: Probabilistic integration framework.}
We do not integrate naively by taking AND or OR operations.
Instead, we probabilistically integrate microscopic (optical flow) and macroscopic (segmentation) perspectives.
This probabilistic combination enables our algorithm to achieve high accuracy robustly.
Unlike end-to-end trained methods such as AdversarialNet, which used test data during training, FoELS maintains consistent parameter settings across all datasets without any dataset-specific training, demonstrating stronger generalization capability.

\textbf{Contribution 3: Original improvement for parallel motion detection.}
Through experimentation, we discovered that probabilistic integration alone is insufficient for parallel motion scenarios, which previous methods also cannot solve.
We therefore developed a novel solution: incorporating optical flow length consideration into the FoE-based likelihood (Challenge \ref{challenge3}), detailed in \ref{subsec:moving_pixel_probability_estimation}.

\section{Related Works}
This section reviews prior efforts in optical flow estimation, segmentation, and moving object detection, which form the foundation of our approach.

\subsection{Optical Flow Estimation}
The field of optical flow estimation has evolved significantly with deep learning approaches.
CNN-based approaches \cite{Dosovitskiy2015FlowNet,Sun2018PWCNet} demonstrated end-to-end optical flow prediction.
RAFT \cite{zachary2020RAFT} introduced 4D correlation volumes and GRU \cite{chung2014gru} refinement, dramatically improving accuracy.
Recent transformer-based approaches \cite{Jiang2021GMA,zhaoyang2022FlowFormer,Shi2022CSFlow,Shi2023VideoFlow} including UniMatch \cite{xu2023unifyingflowstereodepth} and MemFlow \cite{Dong2024MemFlow} achieve state-of-the-art accuracy.

\subsection{Segmentation}
Segmentation has been extensively studied in computer vision \cite{kirillov2023segment,ke2023segmenthighquality,wang2023onepeace,yahuiyuan2020ocr,Zhang2024EfficientViTSAMAS,cai2023efficientvit,Wang_2023_CVPR}.
Promptable segmentation models such as SAM~\cite{kirillov2023segment}, SAM~2~\cite{ravi2024sam2}, and SAM~3~\cite{carion2025sam3} are powerful, but FoELS requires panoptic segmentation so that each segment carries a class label from which we can derive its class-wise moving probability.
Panoptic segmentation, which combines semantic and instance segmentation, is essential for distinguishing between individual object instances.
Among recent models \cite{cheng_masked-attention_2022}, OneFormer \cite{jain2023oneformer} is currently the state-of-the-art panoptic segmentation model.

\subsection{Moving Object Detection}
There are numerous methods for detecting moving objects using static cameras.
For instance, Rozumnyi et al. \cite{Rozumnyi_2021} detect fast-moving objects using static cameras.
However, research on moving object detection from moving cameras remains limited.

Notable review papers on moving object detection include \cite{CHAPEL2020100310}, \cite{Hou2023survey}, and \cite{ZHAO202228}.
Based on these reviews, we have identified several key areas of research in moving object detection, including flow orientation-based methods, focus of expansion (FoE) based approaches, and adversarial network methods.
The FoE is the point from which optical flow vectors radiate when the camera moves forward \cite{Gibson1950}.
Moving objects exhibit flow patterns deviating from this radial structure, making FoE-based analysis effective for motion detection.

Zhao et al. \cite{ZHAO202228} categorized moving object detection application conditions into two types: detection of unseen scenes and detection of seen scenes.
They mainly focused on the latter and background subtraction methods.
However, the paper does not address how to handle moving backgrounds.
MU-Net2 \cite{9413211}, one of the best approaches listed in their survey, is only effective for slightly moving cameras.

There is a similar task for moving object detection, namely ``Semi-Supervised Video Object Segmentation on DAVIS''.
The current best method for this task is HMMN \cite{seong2021hierarchical}.
However, this method requires human initialization, meaning it is not truly moving object detection but rather object tracking.
Therefore, it falls outside the scope of this work.

The best method for DAVIS 2016 listed in the survey by Hou et al.~\cite{Hou2023survey} is AOT~\cite{Yang2021AOT}, and the top-ranked method on the official DAVIS 2016 leaderboard is STM~\cite{Oh2019STM}; however, both are semi-supervised methods requiring human-provided first-frame masks, placing them outside our unsupervised task scope. The DAVIS 2016 leaderboard has not been updated since 2020. Among more recent works, M$^3$Former~\cite{Homeyer2023M3Former} addresses moving object segmentation from a monocular moving camera using a two-stream Mask2Former~\cite{cheng_masked-attention_2022}-based transformer architecture that fuses RGB appearance and motion information via a shared transformer decoder, using frozen off-the-shelf models to generate motion representations---RAFT~\cite{zachary2020RAFT} for optical flow, RAFT-3D~\cite{Teed2021RAFT3D} for scene flow, or higher-dimensional motion costs involving DPT~\cite{Ranftl2021DPT} for depth estimation---alongside ResNet50~\cite{He2016ResNet} backbones. This modular use of off-the-shelf components validates the approach that our method also adopts. However, M$^3$Former does not provide publicly available code nor report IoU on the DAVIS 2016 dataset. Moreover, as a transformer-based architecture, M$^3$Former requires large-scale training data and extensive parameter tuning, whereas FoELS relies on a simple joint probability formulation with far fewer parameters to tune. Our code is publicly available at \url{https://github.com/MasahiroOgawa/foels}.

While not evaluated in the aforementioned review papers, other notable approaches
(\cite{izumiya2002moving,zhangw2020,zhenchenghu1999,zhenchenghu2000,NEGAHDARIPOUR1989303,yanhchaoyang2019})
demonstrate effectiveness for moving object detection.
They can be classified into the following three categories.

\begin{enumerate}
    \item Flow orientation-based approach: \\
          Numerous methods utilize optical flow for moving object detection, such as \cite{izumiya2002moving}.
          Zhang et al. \cite{zhangw2020} introduced a technique that calculates optical flow orientation between adjacent video frames and reconstructs a background orientation field using Poisson fusion.
          This method aims to identify motion saliency by analyzing the discrepancies between the reconstructed background orientation and the observed orientation.
          While it works well for small camera movements, it fails when the camera moves straight ahead, where flow orientations are radially symmetrical.

    \item FoE-based approach: \\
          FoE-based approaches estimate camera motion parameters such as rotation and translation.
          These methods assume a fixed FoE and identify moving objects by analyzing flow vectors relative to the FoE \cite{zhenchenghu1999,zhenchenghu2000}.
          Although conceptually robust, the methods struggle with scenarios involving unknown or dynamic FoE, limiting their utility in real-world conditions.
          While a direct FoE computation method (without optical flow) \cite{NEGAHDARIPOUR1989303} was developed when optical flow was unreliable,
          current optical flow-based FoE estimation is more precise and prevalent, overcoming the limitations of the former
          (e.g., reliance on grayscale images and lack of quantitative evaluation).

    \item Adversarial network approach: \\
          Yang et al. \cite{yanhchaoyang2019} leveraged adversarial learning frameworks to enhance motion detection.
          The generator-inpainter architecture trains the network to distinguish between moving and static regions by minimizing a loss function that encodes flow discrepancies.
          Despite achieving state-of-the-art performance on multiple existing datasets,
          it fails to detect moving objects in low-textured areas and generates false positives for close static objects when tested on our custom traffic video data.
\end{enumerate}

All these representative methods rely primarily on optical flow for moving object detection, making them ineffective in complex, structured scenes.
To address these limitations, we propose a novel approach that integrates both optical flow and segmentation information.

\section{Proposed Method Overview}
\label{section:overview}
This section gives a high-level overview of FoELS; the algorithmic details of each component follow in Section~\ref{section:proposed_method}.

\subsection{System Overview}
Our system overview is illustrated in Fig.~\ref{fig:foels_detail}.
The proposed pipeline consists of six main stages:
\begin{enumerate}
    \item \textbf{Optical Flow Estimation:} Captures pixel-wise motion cues between consecutive frames.
    \item \textbf{Segmentation:} Assigns class-specific prior moving probabilities and identifies static regions.
    \item \textbf{Camera Motion Detection:} Determines if the camera is in motion by analyzing the optical flow ratio within static regions.
    \item \textbf{FoE Computation:} Utilizes Random Sample Consensus (RANSAC) to compute the FoE from optical flow.
    \item \textbf{Moving Pixel Probability Estimation:} An FoE-based moving pixel likelihood is computed from RANSAC outliers. This likelihood is then multiplied by segmentation-derived priors to yield the joint moving pixel probability.
    \item \textbf{Object-Level Refinement:} Validates moving pixel regions against panoptic segmentation results.
\end{enumerate}

\subsection{System Details}
First, frames \( t-1 \) and \( t \) are used to compute the optical flow.
Simultaneously, segmentation is performed on frame \( t \), and each pixel is assigned a prior moving probability according to a manually predefined class-moving probability table.
Sky regions identified through segmentation are removed, since optical flow cannot be computed there.
Concurrently, static areas (e.g., ground, mountains, and buildings) are identified, and the flow within these regions is analyzed.
If the flow existing ratio in the static area exceeds a specified threshold, the camera is considered to be in motion, and the FoE is computed using RANSAC.
The inliers from the RANSAC process are attributed to camera motion, while the outliers are considered as moving pixels.
Once the moving pixels are identified, we map them back to moving objects using the panoptic segmentation results.

\begin{figure}[tbp]
    \vspace{0.5\baselineskip}
    \centering
    \includegraphics[width=0.85\columnwidth]{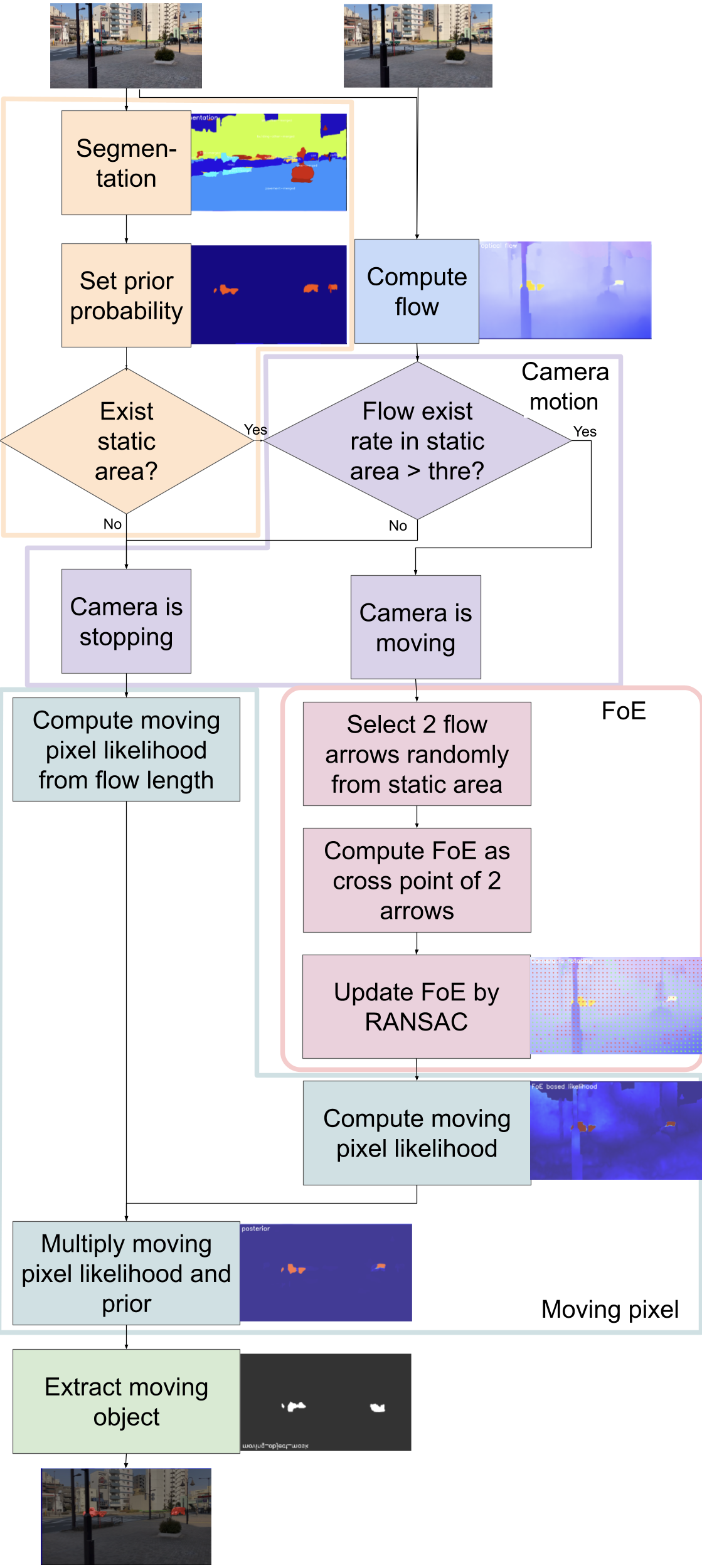}
    \caption{Detailed flowchart of the proposed method.
        The side image illustrates the process outlined in the flowchart.}
    \label{fig:foels_detail}
\end{figure}

The detailed flow chart of the above procedure is listed in Fig.~\ref{fig:foels_detail}.

\section{Proposed Method}
\label{section:proposed_method}
This section details the core algorithmic components introduced in the overview.

\subsection{Optical Flow Estimation}
Based on our research in Section II and Sintel benchmark rankings, we identified UniMatch \cite{xu2023unifyingflowstereodepth} and MemFlow \cite{Dong2024MemFlow} as top candidates with publicly available code.
Quantitative comparisons using the DAVIS 2016 dataset (Table~\ref{table:optical_flow_comparison}) led us to select UniMatch, which provides dense flow maps capturing subtle motion patterns critical for subsequent FoE-based analysis.

\subsection{Segmentation}
\label{subsec:segmentation}
As discussed in Section II, we utilize the OneFormer panoptic segmentation model, which is currently the state-of-the-art approach.
In FoELS, each class is assigned a prior probability reflecting its tendency to be dynamic.
For instance, sky and building classes have low moving probabilities, while vehicle and pedestrian classes have higher values.

In FoELS, these class-based prior moving probabilities are manually defined.
These values were determined and adjusted based on several experiments, and the same predefined values are used for all datasets.
Our class probability table contains 133 predefined classes. For objects classified as ``unknown'' by the segmentation model (i.e., not matching any predefined class), we assign a default moving probability of 0.01.
Although the table covers 133 classes, the prior values themselves come from only a small set of intuitive motion-behavior categories (e.g., likely-moving, movable, might-be-moved, less-moving, never-moving).

It is important to note that the prior is only one factor in the moving pixel probability; even if segmentation is imperfect, the likelihood provides an assessment that can compensate through the multiplicative combination (see Section~\ref{subsec:moving_pixel_probability_estimation}).
Furthermore, segmentation is used to identify static regions for FoE computation. This does not require highly accurate segmentation because ground surfaces are almost always present and reliably detected, and the purpose of static area segmentation is merely to limit the FoE computation to approximately static regions. Even if some static pixels are missed or some non-static pixels are included, the RANSAC-based FoE estimation is inherently robust to such noise. Therefore, the method is robust to imperfect static area segmentation.
FoELS is designed as a modular framework where the segmentation component can be replaced with any state-of-the-art algorithm. As panoptic segmentation methods continue to advance, FoELS will directly benefit from these improvements without requiring changes to the rest of the pipeline.

\subsection{Camera Motion Detection}
\label{subsec:camera_motion_detection}
After obtaining segmentation results and assigning prior moving probabilities,
static areas are identified as regions where the moving probability is below a manually defined threshold.
The camera is considered to be moving if the ratio of existing optical flow in these static areas exceeds a manually defined threshold~(10\% in our implementation).
This work does not compute the camera's egomotion; instead, it only determines whether the camera is moving.
This determination is sufficient for computing the FoE and identifying moving objects.
For pure rotational camera motion where no valid FoE exists, our system detects this condition through the low flow existing rate in static regions and transitions to a STOPPING state, avoiding incorrect FoE computation.
In crowded scenes where the static background area is small, FoE computation uses only the optical flow from static regions identified via the segmentation-based static mask. If the flow existing rate falls below the threshold, the FoE-based angle analysis is skipped, and instead a flow-length-based moving probability is computed for all pixels using the logarithmic scaling of optical flow magnitude. This flow-length-based probability is then multiplied by the prior to yield the moving pixel probability.

\subsection{FoE Computation}
\label{subsec:foe_computation}
To compute the FoE, two optical flow vectors within the identified static area are selected.
An initial FoE candidate is determined as the intersection point of the lines extended from these two flow vectors.
The sign of this FoE candidate (positive for a source of optical flow, negative for a sink) is concurrently determined from the directions of these flow vectors.
Without this sign, objects moving in the opposite direction cannot be correctly identified as moving.
Finally, the RANSAC algorithm is employed to robustly estimate the FoE from the set of available flow vectors in the static regions.
Our RANSAC implementation uses 100 iterations with a requirement of $>$60\% inlier rate and 1° angular tolerance, providing robustness against noisy optical flow measurements.
As shown in Table~\ref{table:optical_flow_comparison}, FoELS demonstrates robustness to optical flow estimator choice, with only 0.037 IoU difference between UniMatch and MemFlow.

\subsection{Moving Pixel Probability Estimation}
\label{subsec:moving_pixel_probability_estimation}
In this work, the moving pixel probability is computed as the joint probability of the segmentation-based moving prior and the FoE-based moving likelihood.

The moving likelihood, computed from the FoE, is multiplied by segmentation-derived priors to yield joint moving pixel probabilities.
This FoE-based moving likelihood is determined from outliers identified during the RANSAC FoE computation.
These outliers correspond to points where the observed optical flow angle deviates from that expected under the computed FoE.

However, relying solely on this angular difference is insufficient for accurately handling scenarios involving motion parallel to the camera,
where the flow direction of the moving object closely aligns with that of the background.
In such a scene, a portion of the moving object's optical flow shares the same direction as the FoE-radial flow of nearby static points and is therefore incorrectly classified as static by an angle-only criterion.
To address this ambiguity, information regarding differences in flow length is incorporated.
Directly multiplying probabilities derived from length differences can induce false positives for nearby static objects, as their flow magnitudes are often large.
Therefore, a logarithmic factor of the length difference is added to the angle-based moving likelihood, and the sum is subsequently clipped to the range [0, 1].
This approach enables FoELS to successfully detect such parallel-moving objects.
The method primarily emphasizes the flow angle difference while also considering significant flow length discrepancies, particularly for detecting parallel motion.

The moving pixel probability is computed as follows:
\begin{equation}
    P_M \propto P(F \cap S) = P(F|S) \cdot P(S),
\end{equation}
where $P_M$ is the joint moving pixel probability,
$P(S)$ is the segmentation-based moving prior probability,
and $P(F|S)$ is the FoE-based moving likelihood conditioned on the segmentation result.
$P(F|S)$ is defined as:
\begin{align}
    P(F|S) & = \text{clip}_{[0,1]}(P_a + \alpha F_l), \label{eq:pfs}
\end{align}
where the function $\text{clip}_{[0,1]}()$ clips a value to the range [0, 1].
$P_a$ is the angle-based probability, which will be detailed below.
$\alpha$ serves as the weighting factor for the flow length,
and $F_l$ is the length factor, also detailed below.

Here, the angle-based probability \( P_a \) is calculated proportionally to the optical flow angle difference
between the optical flow at the point and the expected flow direction based on the FoE.
\( P_a \) is normalized to become 0.5 at a predefined angle difference threshold, \(\theta_{th}\).
The length factor, \(F_l\), incorporates the base-10 logarithm of the flow length difference.
This logarithmic scaling allows for the consideration of significant flow magnitude differences, pertinent for parallel motion, while diminishing the influence of minor variations.
These components are specifically formulated as:
\begin{align}
    P_a & = \text{clip}_{[0,1]}(0.5 \cdot d_a / \theta_{th}),                                                    \\
    F_l & = |\text{log}_{10}(1 + |d_l|)|, \label{eq:fl}                                         \\
    d_a & = \arccos\left(\frac{\mathbf{v}_F \cdot \mathbf{v}_P}{||\mathbf{v}_F|| \cdot ||\mathbf{v}_P||}\right), \\
    d_l & = ||\mathbf{v}_P|| - \overline{||\mathbf{v}_{P,static}||},
\end{align}
where, $\mathbf{v}_F$ is the vector from the FoE to the point,
and $\mathbf{v}_P$ is the optical flow vector at the point.
The term $d_a$ represents the angular difference calculated from these vectors.
$d_l$ is the difference between the pixel's flow magnitude and the mean static-region flow magnitude $\overline{||\mathbf{v}_{P,static}||}$.

All thresholds were empirically determined.
In our experiments, the weighting factor $\alpha$ was set to 0.25, and the angle threshold $\theta_{th}$ to 30~degrees, which is the angle difference at which $P_a$ becomes 0.5 (Eq.~(3)).

In Eq.~(\ref{eq:fl}) the logarithm suppresses the divergence of $F_l$ for very large length differences, preventing nearby static objects with parallax-inflated flow from being misclassified as moving.
The $+1$ inside the logarithm keeps $F_l$ well defined at $|d_l|=0$ and yields the empirical winner among the four $F_l$ variants we evaluated (Table~\ref{table:fl_ablation}).
The remaining thresholds in this paper are dimensionless quantities (angles, probabilities, or ratios) and can be used directly across different resolutions and camera speeds; in our experiments the same parameter set worked unchanged across DAVIS 2016 (480p) and our newly captured datasets ($\sim$1280-wide).

\subsection{Object-Level Refinement}
\label{subsec:object_level_refinement}
Finally, the computed moving pixel probabilities are aggregated to an object level.
This step is crucial for ensuring that an entire object is classified as moving,
even if only a portion of it exhibits detectable motion (addressing Challenge \ref{challenge4}).
Our approach first computes per-pixel probabilities, then integrates them at the segment level.
To achieve this, a binary moving pixel mask is first generated by thresholding the joint moving pixel probability $P_M'$.
A threshold of $0.5^2 = 0.25$ is used for $P_M'$, reflecting the fact that $P_M'$ is a product of two probabilities ($P(S)$ and $P(F|S)$);
this threshold implies that both contributing probabilities are at least 0.5.
This constitutes the first stage of our two-stage filtering process: pixel-level thresholding.
Subsequently, an object-level moving mask is derived.
For each object instance identified by the panoptic segmentation,
the percentage of pixels within that instance that are marked as moving in the binary pixel mask is calculated.
If this percentage exceeds a threshold of 0.01, the entire object instance is classified as moving.
This object-level fraction threshold constitutes the second stage of filtering.
This low threshold is employed to effectively detect objects where only a small part is in motion,
such as the tail of an animal or a limb of a person,
which can sometimes constitute as little as approximately 3\% of the total object area.
The two-stage filtering (pixel-level probability threshold of 0.25 followed by object-level fraction threshold of 0.01) prevents noise from causing false positives: a pixel must have both a high prior and high likelihood to pass Stage 1, and a sufficient fraction of an object's pixels must pass Stage 1 for the object to be classified as moving in Stage 2.

\subsection{Comparison of Tractable Scenes}
To provide a concise comparison of the advantages and disadvantages of related works and the proposed method based on tractable scenes,
we present a comparison table of tractable scenes in Table~\ref{table:tractable_scenes}.
The proposed method, FoELS, is capable of handling a broader range of scenarios compared to existing methods.

\begin{table*}[tbp]
    \vspace{0.5\baselineskip}
    \caption{Comparison of tractable scenes. $\times$: Not tractable, $\triangle$: Partially tractable, \checkmark: Tractable\\
        The possible reasons for tractability are listed in the bottom row for FoELS.}
    \centering
    \resizebox{1.0\textwidth}{!}{
        \begin{tabular}{|c|c|c|c|c|c|c|c|}
            \hline
            \textbf{Method}                        & \textbf{Stop}  & \textbf{Go Forward} & \textbf{Rotate} & \textbf{Go Forward and Rotate} & \textbf{Textureless object} & \textbf{Close object} & \textbf{Close dominant object} \\ \hline
            Flow Orientation \cite{zhangw2020}     & \checkmark     & {$\times$}          & {$\times$}      & {$\times$}                     & {$\times$}                  & \checkmark            & {$\times$}                     \\ \hline
            FoE \cite{zhenchenghu2000}             & \checkmark     & \checkmark          & {$\times$}      & {$\times$}                     & {$\times$}                  & \checkmark            & {$\times$}                     \\ \hline
            AdversarialNet \cite{yanhchaoyang2019} & \checkmark     & \checkmark          & {$\triangle$}   & {$\triangle$}                  & {$\times$}                  & {$\times$}            & {$\times$}                     \\ \hline
            \rowcolor{gray!20}
            FoELS (Ours)                           & \checkmark     & \checkmark          & {$\triangle$}   & {$\triangle$}                  & \checkmark                  & \checkmark            & {$\times$}                     \\
                                                   & by Orientation & by FoE              &                 &                                & by Seg                      & by FoE                &                                \\ \hline
        \end{tabular}
    } 
    \label{table:tractable_scenes}
\end{table*}

\section{Evaluation}
\subsection{Datasets}
\label{subsec:datasets}
Experiments were conducted on the DAVIS 2016 dataset \cite{Perazzi2016}, the FBMS-59 dataset \cite{OB14b}, and a custom-collected traffic video dataset.
The DAVIS 2016 and FBMS-59 datasets, which are annotated for moving objects, were utilized for quantitative evaluation.
These relatively small datasets pose a risk of overfitting for training-based approaches.
Though our method involves fitting only a few parameters, rather than comprehensive training,
this risk is pertinent to the training-dependent methods against which we compare.
To evaluate robustness and applicability in real-world scenarios,
the custom traffic video dataset, which is unannotated, was used for qualitative assessment.

We selected DAVIS 2016 and FBMS-59 specifically for fair comparison with AdversarialNet, which published results on these same datasets. DAVIS 2017 provides semi-supervised and unsupervised tasks. The semi-supervised task requires human initialization (first-frame mask), which is essentially object tracking rather than autonomous moving object detection. The unsupervised task is defined as segmenting objects that capture human attention (saliency-based), which is closer to salient object segmentation than to moving object detection. Both tasks thus fall outside our problem scope. CDNet primarily focuses on background subtraction for surveillance with mostly static cameras, targeting different application scenarios. Our custom traffic dataset addresses real-world scenarios not covered by standard benchmarks.

\subsubsection{Quantitative Evaluation Dataset}
For quantitative evaluation, we utilized the DAVIS 2016 dataset and the FBMS-59 dataset.

The FBMS-59 dataset provides annotations specifically for the moving object detection task.
In contrast, the DAVIS 2016 dataset is primarily designed for video object segmentation,
which is a binary labeling problem focused on separating foreground objects from the background in a video.
Consequently, the foreground annotations in DAVIS 2016 may sometimes include objects that are part of a moving background.

Upon careful examination of the DAVIS 2016 dataset,
it was observed that certain scenes are inappropriate for evaluating moving object detection due to the presence of unannotated moving backgrounds.
For instance, the \texttt{breakdance} scene features background spectators in motion who are not labeled as moving objects.
The dataset comprises 50 scenes in total.
After identifying and excluding scenes with significant unannotated background motion,
the following three scenes were removed: {\small \texttt{bmx-bumps}, \texttt{breakdance}, and \texttt{dance-jump}} (3 out of 50).

Furthermore, an additional 15 scenes exhibit slight, unannotated background motion.
However the background movements in these scenes are minor, and to maintain a substantial dataset size for evaluation,
they were retained in our evaluation set.
Consequently, the final evaluation set, termed DAVIS 2016 train-val-movobj, consists of the remaining 47 scenes.

\subsubsection{Qualitative Evaluation Dataset}
To assess the performance of FoELS in real-world conditions, a custom traffic video dataset was captured.
This dataset was specifically designed to address failure modes of existing methods and evaluate FoELS's contributions.
The chosen scenarios are:
(1) \textbf{Parallel-moving vehicles}: The most challenging scenario where the flow direction of moving objects closely aligns with the background, requiring our novel flow length consideration approach.
(2) \textbf{Stationary vehicles}: Common in traffic scenarios, testing the ability to distinguish truly moving objects from temporarily stopped ones.
(3) \textbf{Low-textured environments}: Challenging for optical flow-based methods, demonstrating the value of integrating segmentation as a macroscopic cue.
(4) \textbf{Camera zoom}: Not represented in standard datasets (DAVIS 2016, FBMS-59), testing the robustness of FoE-based approach to non-translational camera motion.
Additionally, we added experiments on opposite-direction, cross-direction motion, and crowded scenes to demonstrate FoELS's general applicability beyond these specific scenarios.

\subsection{Evaluation Metrics}
We adopt Intersection-over-Union (IoU) scores as the primary evaluation metric,
consistent with the methodology employed by the Adversarial Network \cite{yanhchaoyang2019}.
This facilitates a direct comparison of FoELS's performance against that of the Adversarial Network.
The scene IoU score is calculated by averaging the IoU scores across all frames within a sequence.
The final IoU score is subsequently determined by averaging all computed scene IoU scores.

\subsection{Results}
\label{subsec:results}
\begin{table}[tbp]
    \caption{Quantitative evaluation result.
        The values represent the average IoU scores over the DAVIS 2016 train-val-movobj sequences and FBMS-59 Testset scenes.}
    \centering
    \begin{tabular}{|l|c|c|}
        \hline
        \textbf{Method} & DAVIS 2016     & FBMS 59        \\
        \hline
        Adversarial Net & 0.599          & 0.369          \\
        \hline
        FoELS (Ours)    & \textbf{0.773} & \textbf{0.695} \\
        \hline
    \end{tabular}
    \label{table:results}
\end{table}

The final quantitative evaluation results are presented in Table~\ref{table:results}.
The Adversarial Network's training protocol included the use of test data.
In contrast, FoELS was trained without access to test data and employed consistent settings across all datasets.
Despite this difference in training methodology, FoELS surpassed the state-of-the-art Adversarial Network method, achieving a higher IoU score.

\begin{figure}[tbp]
    \vspace{0.5\baselineskip}
    \centering
    \includegraphics[width=0.8\columnwidth]{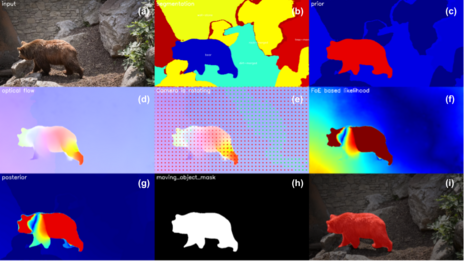}
    \caption{Example visual results of FoELS on the DAVIS 2016 bear scene.
        \textbf{First row (left to right):} (a) Input frame, (b) segmentation result, and (c) prior moving probability derived from segmentation.
        \textbf{Second row (left to right):} (d) Optical flow, (e) optical flow with FoE inlier (green arrows) and outliers (red arrows), and (f) the FoE-based moving likelihood.
        \textbf{Third row (left to right):} (g) Joint moving pixel probability, (h) refined object-level moving mask, and (i) the final moving object result.}
    \label{fig:comb_bear}
\end{figure}

Fig.~\ref{fig:comb_bear} shows an example of the visual results from the above evaluation.
This figure illustrates the step-by-step results of the process detailed in Section~\ref{section:proposed_method}.
The first row displays: (a) the input frame, (b) the segmentation result, where different colors denote distinct classes,
and (c) the prior moving probability derived from segmentation.
The prior probability is visualized using a jet colormap, where red indicates higher probability and blue signifies lower probability.
The second row presents: (d) the optical flow, with orientation encoded by color,
(e) the optical flow field highlighting FoE inliers (green arrows) and outliers (red arrows).
An existing FoE in the image is marked with a thick red cross.
(f) The FoE-based moving likelihood, also depicted using a jet colormap.
The third row shows: (g) the joint moving pixel probability,
calculated as the product of the prior moving probability and the FoE-based moving likelihood,
(h) the refined object-level moving mask, demonstrating the aggregation of moving pixels to an object level,
and (i) the final moving object mask overlaid on the input image.
In this particular example, the bear's hand remains stationary while the bear is walking.
Nevertheless, FoELS successfully extracts the entire bear due to the object-level refinement process.

\begin{figure}[tbp]
    \centering
    \begin{subfigure}[b]{\columnwidth}
        \includegraphics[width=\columnwidth]{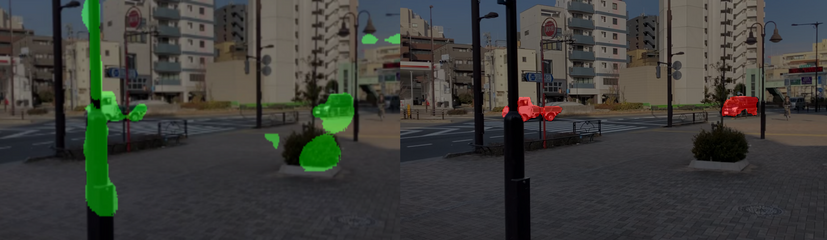}
        \caption{Traffic}\label{fig:todaiura_1_foelsadv_comp}
    \end{subfigure} \\
    \begin{subfigure}[b]{\columnwidth}
        \includegraphics[width=\columnwidth]{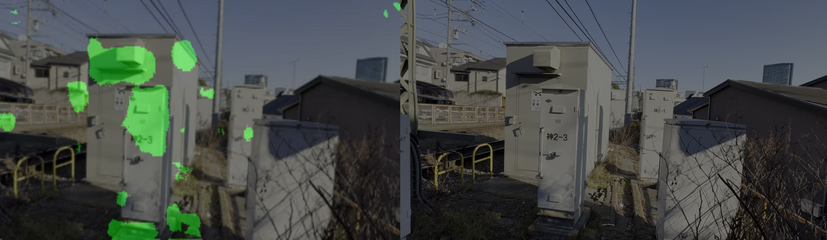}
        \caption{Zoom:initial}\label{fig:komabatrain_1_foelsadv_comp}
    \end{subfigure} \\
    \begin{subfigure}[b]{\columnwidth}
        \includegraphics[width=\columnwidth]{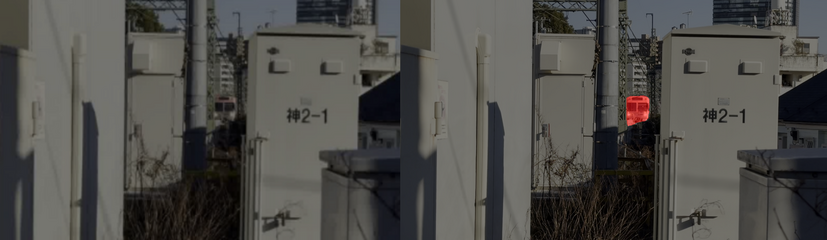}
        \caption{Zoom:train}\label{fig:komabatrain_130_foelsadv_comp}
    \end{subfigure} \\
    \caption{Comparison results with AdversarialNet (left) and FoELS (right) across different scenarios.
        AdversarialNet exhibits limited generalization to unseen scenes, while FoELS maintains robust performance without scene-specific tuning.
        The dramatic visual improvement reflects the difference between real-world complexity and standard datasets.}
    \label{fig:foelsadv_comparisons}
\end{figure}

Fig.~\ref{fig:foelsadv_comparisons} compares the results of the Adversarial Network with those of FoELS across different scenarios.
Fig.~\ref{fig:todaiura_1_foelsadv_comp} shows the traffic scene comparison for the same scene as Fig.~\ref{fig:result_sample}.
The left side shows the results of the Adversarial Network, while the right side displays the results of FoELS.
The Adversarial Network falsely detects nearby vegetation and poles as moving objects due to their significantly different optical flow compared to the background.
In contrast, FoELS successfully identifies only the genuinely moving objects by primarily relying on FoE-based flow orientation analysis.

\begin{figure*}[tbp]
    \vspace{0.5\baselineskip}
    \centering
    \begin{subfigure}[b]{0.33\textwidth}
        \includegraphics[width=\textwidth]{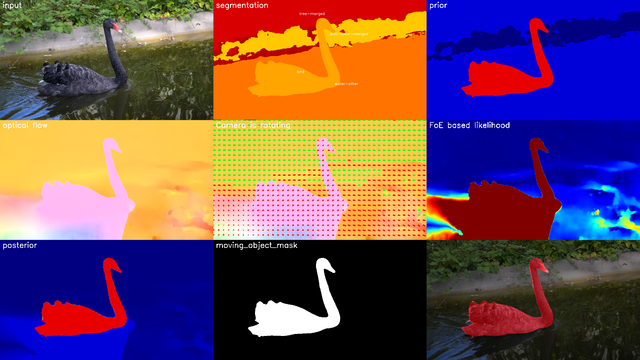}
        \caption{DAVIS:blackswan}\label{fig:blackswan_00000_result_comb}
    \end{subfigure}%
    \begin{subfigure}[b]{0.33\textwidth}
        \includegraphics[width=\textwidth]{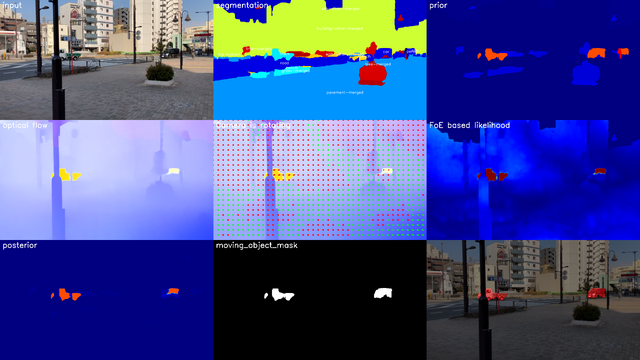}
        \caption{Custom:traffic}\label{fig:comb_todaiura1}
    \end{subfigure}%
    \begin{subfigure}[b]{0.33\textwidth}
        \includegraphics[width=\textwidth]{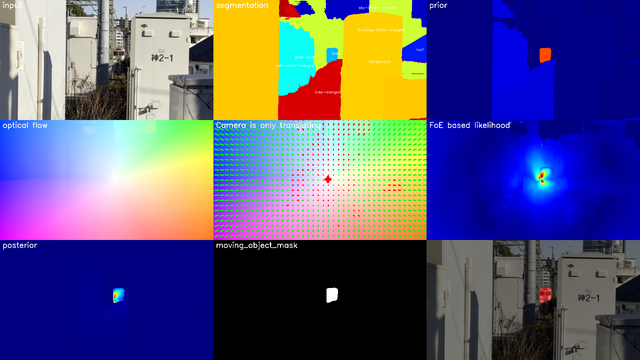}
        \caption{Custom:zoom}\label{fig:comb_komabatrain130}
    \end{subfigure} \\
    \begin{subfigure}[b]{0.33\textwidth}
        \includegraphics[width=\textwidth]{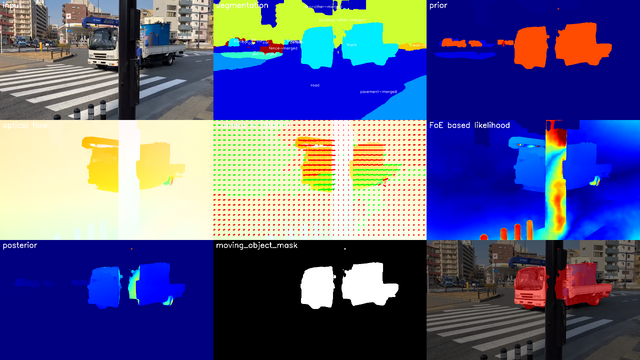}
        \caption{Custom:parallel}\label{fig:todaiura_441_result_comb}
    \end{subfigure}%
    \begin{subfigure}[b]{0.33\textwidth}
        \includegraphics[width=\textwidth]{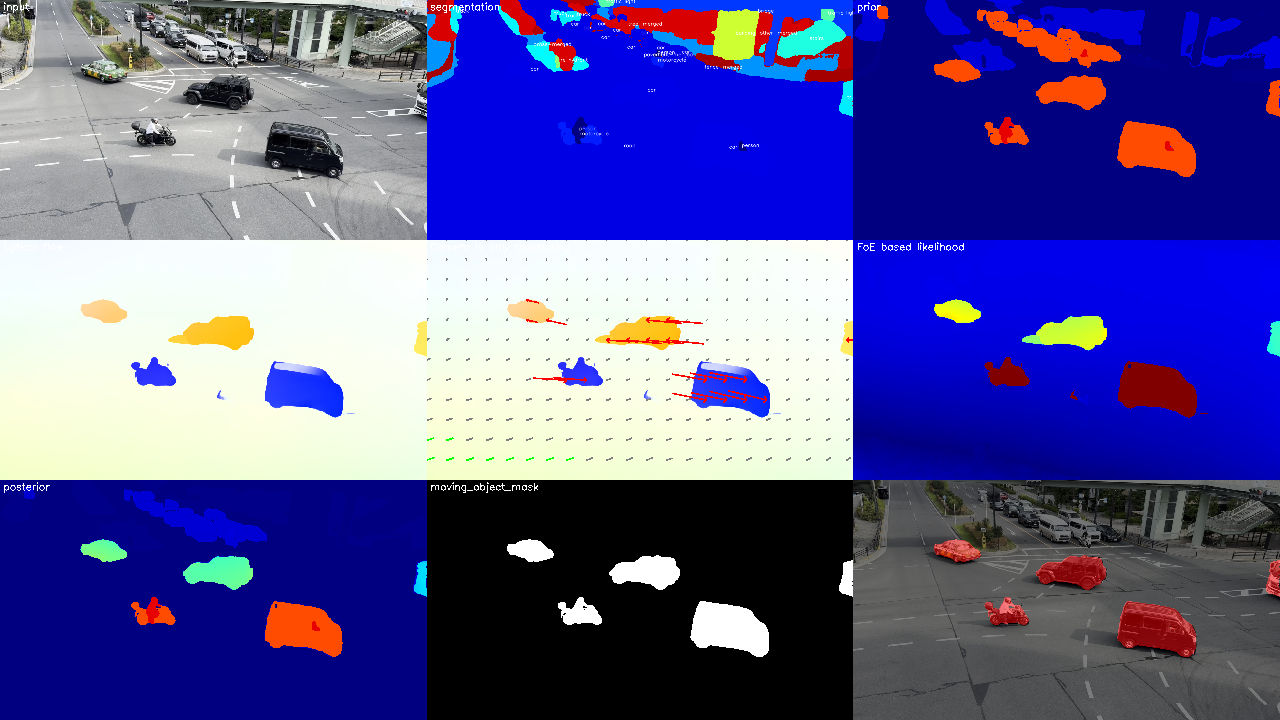}
        \caption{Custom:cross}
    \end{subfigure}%
    \begin{subfigure}[b]{0.33\textwidth}
        \includegraphics[width=\textwidth]{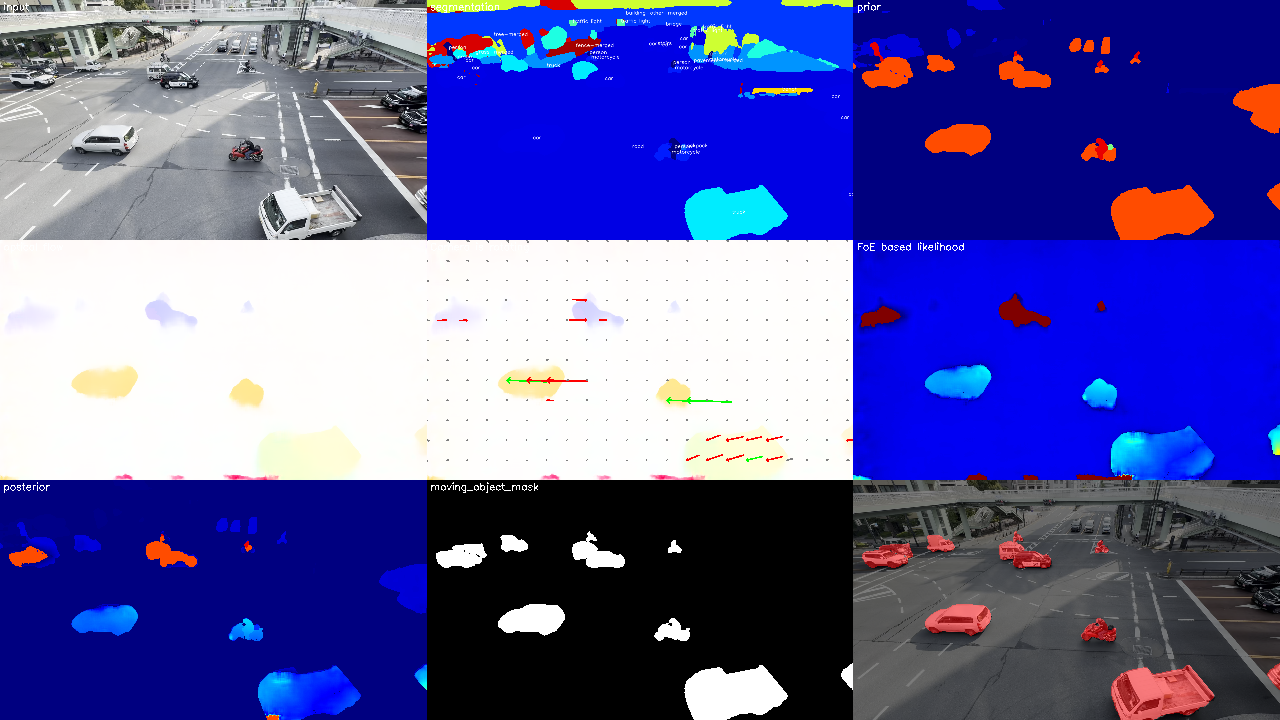}
        \caption{Custom:opposite in crowded}
    \end{subfigure}
    \caption{Example visual results of FoELS on various motion types including parallel, opposite-direction, cross-direction, and crowded scenes.
        See Fig.~\ref{fig:comb_bear} for the 9-subimage format.}
    \label{fig:FoELS_results}
\end{figure*}

Fig.~\ref{fig:comb_todaiura1} shows the step-by-step visualization results of the same scene as Fig.~\ref{fig:result_sample}.
In this example, it can be seen why FoELS can correctly detect cars almost moving parallel to the camera,
and the nearby static pole, despite exhibiting large optical flow, is correctly identified as stationary.

Fig.~\ref{fig:komabatrain_1_foelsadv_comp} compares the results of the Adversarial Network with those of FoELS
on the custom zoom-in/zoom-out video evaluation, where the camera remains stationary while zooming.
The left panel illustrates the results of the Adversarial Network, while the right panel displays the results of FoELS.
In this initial frame, the camera is nearly stationary, and no zoom is applied.
However, the Adversarial Network produces numerous false positives.
This indicates a lack of robustness in the Adversarial Network when applied to novel, untrained scenes.
Conversely, FoELS exhibits no false positives in this scenario.

Fig.~\ref{fig:komabatrain_130_foelsadv_comp} presents a similar comparison between the Adversarial Network and FoELS.
In this instance, the camera is actively zooming in while the train is in motion.
An incoming train is positioned near the center of the image,
while simultaneously the background exhibits motion due to the camera zoom.
Notably, the Adversarial Network fails to detect any moving objects.
In contrast, FoELS successfully identifies the approaching train while correctly disregarding the background motion induced by the zoom.

Fig.~\ref{fig:comb_komabatrain130} illustrates the intermediate processing steps for the same frame of Fig.~\ref{fig:komabatrain_130_foelsadv_comp},
employing the visualization format detailed in Fig.~\ref{fig:comb_bear}.
This visualization demonstrates the successful detection of the moving train by FoELS.

We present additional visual results of FoELS in Fig.~\ref{fig:FoELS_results}.
Fig.~\ref{fig:blackswan_00000_result_comb} shows an example of the black swan scene in the DAVIS 2016 dataset.
The swan's color is very close to the background river, making it hard to segment the swan.
Prior to selecting the final segmentation model,
we evaluated several state-of-the-art approaches and found that OneFormer \cite{jain2023oneformer},
the model ultimately adopted, successfully segments the swan,
thereby enabling FoELS to detect its motion even in this challenging scene.

These scenes contain challenging scenarios, where some potential moving objects are in motion while others remain static.
However, FoELS successfully detects the moving objects in all of them.

\subsection{Ablation Studies}
We conducted ablation studies to evaluate the effectiveness of each component of FoELS.
The results are presented in Table~\ref{table:ablation_studies}.
The study began with a comparison of semantic and panoptic segmentation.
Specifically, we evaluated the semantic segmentation models InternImageT,
as well as the panoptic segmentation model OneFormer.
For the OneFormer model, object refinement was subsequently incorporated.
Finally, the inclusion of the FoE sign and further parameter adjustments constituted the final FoELS configuration.

We also compared different optical flow methods to select the most suitable one for FoELS.
Table~\ref{table:optical_flow_comparison} shows the quantitative comparison between MemFlow and UniMatch.

We also report a four-way ablation of the length-factor formula $F_l$ in Table~\ref{table:fl_ablation}.
The current-code formulation \texttt{diff\_log} ($F_l=|\log_{10}(1+|\Delta\text{length}|)|$) achieves the highest macro-mean IoU, validating the saturating log compression discussed in Subsection~\ref{subsec:moving_pixel_probability_estimation}; the spread across the four variants is only 0.009 IoU, confirming that the formulation is robust to the choice of $F_l$.
We additionally report a sensitivity sweep over $\alpha$ in Table~\ref{table:alpha_sweep}.
The chosen value $\alpha=0.25$ has the highest macro-mean IoU, but its margin over $\alpha=0.5$ is only 0.005 IoU, well within the noise of this benchmark; the formulation is therefore robust to $\alpha$ within $[0.25, 0.5]$, and even $\alpha=0$ (parallel-motion correction disabled) only loses 0.026 IoU.

\begin{table}[tbp]
    \caption{Ablation study results.
        The values represent the average IoU scores over the DAVIS 2016 train-val-movobj sequences.
        "OneFormer with ObjRefine" refers to the OneFormer model for panoptic segmentation with object refinement.
        "+ FoE sign" indicates the addition of the FoE sign, representing the final FoELS configuration.}
    \centering
    \begin{tabular}{|l|c|}
        \hline
        \textbf{Method}                     & IoU            \\
        \hline
        InternImageT (Semantic)             & 0.532          \\
        \hline
        Oneformer (Panoptic) with ObjRefine & 0.65           \\
        \hline
        + FoE sign (=FoELS)                 & \textbf{0.773} \\
        \hline
    \end{tabular}
    \label{table:ablation_studies}
\end{table}

\begin{table}[tbp]
    \caption{Comparison of optical flow methods.
        The values represent the average IoU scores of FoELS over the DAVIS 2016 train-val-movobj sequences using different optical flow methods.}
    \centering
    \begin{tabular}{|l|c|}
        \hline
        \textbf{Optical Flow Method} & IoU            \\
        \hline
        MemFlow                      & 0.736          \\
        \hline
        UniMatch (FoELS)             & \textbf{0.773} \\
        \hline
    \end{tabular}
    \label{table:optical_flow_comparison}
\end{table}

\begin{table}[tbp]
    \caption{Ablation of the length-factor formula $F_l$ in Eq.~(\ref{eq:fl}) on the DAVIS 2016 train-val-movobj benchmark (47 sequences, 3{,}508 frames), with $\alpha=0.25$ and all other parameters fixed.
        Let $\Delta\ell=\|\mathbf{v}_P\|-\overline{\|\mathbf{v}_{P,\text{static}}\|}$ and $d_l^{\text{ratio}}=\|\mathbf{v}_P\|/\overline{\|\mathbf{v}_{P,\text{static}}\|}$.
        \texttt{diff\_log} is the formula used in the paper (Eq.~(\ref{eq:fl})).}
    \centering
    
    \resizebox{\columnwidth}{!}{%
    \begin{tabular}{|l|c|c|c|}
        \hline
        \textbf{Formula} & \textbf{$F_l$} & \textbf{Macro-mean IoU} & \textbf{Std} \\
        \hline
        \texttt{diff\_log} (FoELS) & $|\log_{10}(1+|\Delta\ell|)|$ & \textbf{0.773} & 0.183 \\
        \hline
        \texttt{diff\_lin}         & $|\Delta\ell|$               & 0.764          & 0.195 \\
        \hline
        \texttt{ratio\_log}        & $|\log_{10}(d_l^{\text{ratio}})|$ & 0.765          & 0.186 \\
        \hline
        \texttt{ratio\_lin}        & $|d_l^{\text{ratio}}-1|$          & 0.769          & 0.182 \\
        \hline
    \end{tabular}}
    \label{table:fl_ablation}
\end{table}

\begin{table}[tbp]
    \caption{Sensitivity sweep over the length-factor weight $\alpha$ in Eq.~(\ref{eq:pfs}) on the DAVIS 2016 train-val-movobj benchmark (47 sequences, 3{,}508 frames).
        Macro-mean IoU is computed by averaging per-sequence mean IoUs across the 47 sequences; the between-sequence standard deviation is reported alongside.
        $\alpha=0.25$ is the value used in the paper.}
    \centering
    
    \begin{tabular}{|c|c|c|}
        \hline
        \textbf{$\alpha$} & \textbf{Macro-mean IoU} & \textbf{Std} \\
        \hline
        0.00              & 0.746                   & 0.184 \\
        \hline
        0.25 (FoELS)      & \textbf{0.773}          & 0.184 \\
        \hline
        0.50              & 0.767                   & 0.188 \\
        \hline
    \end{tabular}
    \label{table:alpha_sweep}
\end{table}

FoELS currently focuses on offline use and runs at about 5~seconds per frame on an Intel Core~i9-13900HX CPU with an NVIDIA GeForce RTX~4080 GPU; vectorising the per-pixel evaluation is expected to reduce this substantially. The main computational bottleneck is the per-pixel FoE-based moving-probability evaluation, which accounts for over 90\% of the total runtime.

\section{Conclusion}
FoELS is an innovative method for detecting moving objects from a moving camera,
seamlessly integrating optical flow, segmentation, and camera motion detection through FoE estimation.
By addressing challenges such as rotational motion and low-textured environments,
FoELS demonstrates robust performance across diverse scenarios.
Owing to its FoE-centered flow analysis, FoELS can detect objects even during camera zoom operations,
a scenario often challenging for existing moving object detection techniques.
FoELS demonstrates robust performance on the DAVIS 2016 and FBMS-59 datasets, as well as real-world traffic videos,
employing consistent settings across all datasets, underscoring its potential for various applications in robotics and computer vision.
Furthermore, the modular architecture of FoELS, which is not tightly coupled with specific segmentation or optical flow methods,
allows for the integration of future advancements in these areas, potentially leading to further performance enhancements.

FoELS achieves high accuracy but is computationally intensive due to the trade-off between accuracy and real-time feasibility.
Temporal instability also occurs from frame-by-frame processing, which can cause flickering in the detection results.
Future work includes optimization strategies: model pruning, quantization, lighter architectures, parallel processing, and adding tracking.

We acknowledge several limitations of FoELS: (1) Dependence on segmentation quality---while the likelihood can partially compensate for segmentation errors, overall performance is influenced by segmentation accuracy, particularly for novel object categories. (2) Temporal flickering---frame-by-frame processing can cause inconsistent detection across frames; adding temporal tracking would address this. (3) Dominant moving foreground---when moving objects occupy most of the image, leaving insufficient static background for FoE computation, detection accuracy degrades. (4) Fast camera motion and motion blur reduce accuracy, but this affects any image-based method.


\begin{profile}
    \Name{Masahiro Ogawa}
    \Orcid{0009-0005-5820-0026}
    \Affiliation{The University of Tokyo}
    \Address{Department of Precision Engineering, Graduate School of Engineering, The University of Tokyo}
    \History{received the B.S. degree in mathematics and the M.S. degree in cosmology from Kyoto University, Kyoto, Japan, in 2002 and 2005, respectively. From 2005 to 2018, he worked at the Higashifuji Research Laboratory, Toyota Motor Corporation. Since 2020, he has been working at Sensyn Robotics Inc. Since 2022, he has been a Ph.D. student at The University of Tokyo, Tokyo, Japan.}
\end{profile}

\begin{profile}
    \Name{Qi An}
    \Orcid{0000-0001-7641-2632}
    \Affiliation{The University of Tokyo}
    \Address{Department of Human and Engineered Environmental Studies, Graduate School of Frontier Sciences, The University of Tokyo}
    \History{received his B.E., M.E., and Ph.D. degrees in Engineering from The University of Tokyo in 2009, 2011, and 2014, respectively. From 2012, he was a JSPS Research Fellow for Young Scientists (DC1). From 2015 to 2020, he was an Assistant Professor at The University of Tokyo. From 2020 to 2022, he was an Associate Professor at Kyushu University. Since 2022, he has been an Associate Professor at The University of Tokyo.}
    \Membership{$\bullet$ IEEE $\bullet$ JSPE $\bullet$ RSJ $\bullet$ JSME $\bullet$ SICE $\bullet$ LSE}
\end{profile}

\begin{profile}
    \Name{Atsushi Yamashita}
    \Orcid{0000-0003-1280-069X}
    \Affiliation{The University of Tokyo}
    \Address{Department of Human and Engineered Environmental Studies, Graduate School of Frontier Sciences, The University of Tokyo}
    \History{received B.E., M.E., and Ph.D. degrees from the Faculty of Engineering, the University of Tokyo in 1996, 1998, and 2001, respectively. From 2001 to 2008, he was an Assistant Professor of Shizuoka University. From 2006 to 2007, he was a Visiting Associate of California Institute of Technology. From 2008 to 2011, he was an Associate Professor of Shizuoka University. From 2011 to 2022, he was an Associate Professor of The University of Tokyo. Since 2022, he has been a Professor of The University of Tokyo. His research interests are robot vision, image processing, and intelligent robot systems.}
    \Membership{$\bullet$ Fellow of RSJ $\bullet$ ACM $\bullet$ IEEE $\bullet$ JSPE $\bullet$ IEICE $\bullet$ JSAE $\bullet$ JSCE $\bullet$ JSME $\bullet$ IEEJ $\bullet$ IPSJ $\bullet$ ITE $\bullet$ SICE $\bullet$ Society for Serviceology}
\end{profile}


\begin{thebibliography}{10}
\providecommand{\url}[1]{#1}
\csname url@samestyle\endcsname
\providecommand{\newblock}{\relax}
\providecommand{\bibinfo}[2]{#2}
\providecommand{\BIBentrySTDinterwordspacing}{\spaceskip=0pt\relax}
\providecommand{\BIBentryALTinterwordstretchfactor}{4}
\providecommand{\BIBentryALTinterwordspacing}{\spaceskip=\fontdimen2\font plus
\BIBentryALTinterwordstretchfactor\fontdimen3\font minus
  \fontdimen4\font\relax}
\providecommand{\BIBforeignlanguage}[2]{{%
\expandafter\ifx\csname l@#1\endcsname\relax
\typeout{** WARNING: fujipressbib.bst: No hyphenation pattern has been}%
\typeout{** loaded for the language `#1'. Using the pattern for}%
\typeout{** the default language instead.}%
\else
\language=\csname l@#1\endcsname
\fi
#2}}
\providecommand{\BIBdecl}{\relax}
\BIBdecl

\bibitem{zhangw2020}
W.~Zhang, X.~Sun, and Q.~Yu, ``Moving object detection under a moving camera
  via background orientation reconstruction.'' Sensors, Vol.20, No.~3103,
  2020. https://doi.org/\url{10.3390/s20113103}

\bibitem{zhenchenghu1999}
Z.~Hu, K.~Uchimura, and S.~Kawaji, ``Determining motion parameters for
  vehicle-mounted camera using focus of expansion,'' IEEJ Transactions on
  Industry Applications, Vol.~119, No.1, pp.~50-57,  1999.
  https://doi.org/\url{10.1541/ieejias.119.50}

\bibitem{yanhchaoyang2019}
Y.~Yang, A.~Loquercio, D.~Scaramuzza, and S.~Soatto, ``Unsupervised moving
  object detection via contextual information separation,'' Proceedings of the
  IEEE/CVF Conference on Computer Vision and Pattern Recognition (CVPR),
  pp.~879-888,  2019. https://doi.org/\url{10.1109/CVPR.2019.00097}

\bibitem{zhenchenghu2000}
Z.~Hu and K.~Uchimura, ``Multiple moving objects detection and simultaneous
  tracking from the time-varied background,'' IEEJ Transactions on Industry
  Applications, Vol.~120, No.10, pp.~1134-1142,  2000.
  https://doi.org/\url{10.1541/ieejias.120.1134}

\bibitem{Dosovitskiy2015FlowNet}
A.~Dosovitskiy, P.~Fischer, E.~Ilg, P.~H{\"a}usser, C.~Haz{\i}rbas, V.~Golkov,
  P.~van~der Smagt, D.~Cremers, and T.~Brox, ``Flownet: Learning optical flow
  with convolutional networks,''   Proceedings of the IEEE International
  Conference on Computer Vision (ICCV), pp.~2758-2766,  2015.
  https://doi.org/\url{10.1109/ICCV.2015.316}

\bibitem{Sun2018PWCNet}
D.~Sun, X.~Yang, M.-Y. Liu, and J.~Kautz, ``Pwc-net: Cnns for optical flow
  using pyramid, warping, and cost volume,''   Proceedings of the IEEE/CVF
  Conference on Computer Vision and Pattern Recognition (CVPR), pp.~8934-8943,
  2018. https://doi.org/\url{10.1109/CVPR.2018.00931}

\bibitem{zachary2020RAFT}
Z.~Teed and J.~Deng, ``Raft: Recurrent all-pairs field transforms for optical
  flow,''   Proceedings of the European Conference on Computer Vision (ECCV),
  p.~402–419,  2020. https://doi.org/\url{10.1007/978-3-030-58536-5_24}

\bibitem{chung2014gru}
J.~Chung, C.~Gulcehre, K.~Cho, and Y.~Bengio, ``Empirical evaluation of gated
  recurrent neural networks on sequence modeling,''   NIPS 2014 Deep Learning
  and Representation Learning Workshop,  2014.

\bibitem{Jiang2021GMA}
S.~Jiang, D.~Campbell, Y.~Lu, H.~Li, and R.~Hartley, ``Learning to estimate
  hidden motions with global motion aggregation,''   Proceedings of the
  IEEE/CVF International Conference on Computer Vision (ICCV), pp.~9752-9761,
  2021. https://doi.org/\url{10.1109/ICCV48922.2021.00963}

\bibitem{zhaoyang2022FlowFormer}
Z.~Huang, X.~Shi, C.~Zhang, Q.~Wang, K.~C. Cheung, H.~Qin, J.~Dai, and H.~Li,
  ``Flowformer: A transformer architecture for optical flow,''   Proceedings of
  the European Conference on Computer Vision (ECCV), p.~668–685,  2022.
  https://doi.org/\url{10.1007/978-3-031-19790-1_40}

\bibitem{Shi2022CSFlow}
H.~Shi, Y.~Zhou, K.~Yang, X.~Yin, and K.~Wang, ``Csflow: Learning optical flow
  via cross strip correlation for autonomous driving,''   2022 IEEE Intelligent
  Vehicles Symposium (IV), p.~1851–1858,  2022.
  https://doi.org/\url{10.1109/IV51971.2022.9827341}

\bibitem{Shi2023VideoFlow}
X.~Shi, Z.~Huang, W.~Bian, D.~Li, M.~Zhang, K.~C. Cheung, S.~See, H.~Qin,
  J.~Dai, and H.~Li, ``Videoflow: Exploiting temporal cues for multi-frame
  optical flow estimation,''   Proceedings of the IEEE/CVF International
  Conference on Computer Vision (ICCV), pp.~12\,435-12\,446,  2023.
  https://doi.org/\url{10.1109/ICCV51070.2023.01146}

\bibitem{xu2023unifyingflowstereodepth}
H.~Xu, J.~Zhang, J.~Cai, H.~Rezatofighi, F.~Yu, D.~Tao, and A.~Geiger,
  ``Unifying flow, stereo and depth estimation,'' IEEE Transactions on Pattern
  Analysis and Machine Intelligence, Vol.45, No.11, pp.~13\,941-13\,958,  2023.
  https://doi.org/\url{10.1109/TPAMI.2023.3298645}

\bibitem{Dong2024MemFlow}
Q.~Dong and Y.~Fu, ``Memflow: Optical flow estimation and prediction with
  memory,''   Proceedings of the IEEE/CVF Conference on Computer Vision and
  Pattern Recognition (CVPR),  2024.

\bibitem{kirillov2023segment}
A.~Kirillov, E.~Mintun, N.~Ravi, H.~Mao, C.~Rolland, L.~Gustafson, T.~Xiao,
  S.~Whitehead, A.~C. Berg, W.-Y. Lo, P.~Dollár, and R.~Girshick, ``Segment
  anything,''   Proceedings of the IEEE/CVF International Conference on
  Computer Vision (ICCV), pp.~3992-4003,  2023.
  https://doi.org/\url{10.1109/ICCV51070.2023.00371}

\bibitem{ke2023segmenthighquality}
L.~Ke, M.~Ye, M.~Danelljan, Y.~Liu, Y.-W. Tai, C.-K. Tang, and F.~Yu, ``Segment
  anything in high quality,''   Advances in Neural Information Processing
  Systems (NeurIPS),  2023.

\bibitem{wang2023onepeace}
P.~Wang, S.~Wang, J.~Lin, S.~Bai, X.~Zhou, J.~Zhou, X.~Wang, and C.~Zhou,
  ``One-peace: Exploring one general representation model toward unlimited
  modalities,'' arXiv preprint arXiv:2305.11172,  2023.

\bibitem{yahuiyuan2020ocr}
Y.~Yuan, X.~Chen, and J.~Wang, ``Object-contextual representations for semantic
  segmentation,''   Proceedings of the European Conference on Computer Vision
  (ECCV), ser. Lecture Notes in Computer Science, Vol.~12351, pp.~173-190,
  2020. https://doi.org/\url{10.1007/978-3-030-58539-6\_11}

\bibitem{Zhang2024EfficientViTSAMAS}
Z.~Zhang, H.~Cai, and S.~Han, ``Efficientvit-sam: Accelerated segment anything
  model without performance loss,''   Proceedings of the IEEE/CVF Conference on
  Computer Vision and Pattern Recognition (CVPR) Workshops, pp.~7859-7863,
  2024.

\bibitem{cai2023efficientvit}
H.~Cai, J.~Li, M.~Hu, C.~Gan, and S.~Han, ``Efficientvit: Lightweight
  multi-scale attention for high-resolution dense prediction,''   Proceedings
  of the IEEE/CVF International Conference on Computer Vision (ICCV),
  pp.~17\,302-17\,313,  2023.

\bibitem{Wang_2023_CVPR}
W.~Wang, J.~Dai, Z.~Chen, Z.~Huang, Z.~Li, X.~Zhu, X.~Hu, T.~Lu, L.~Lu, H.~Li,
  X.~Wang, and Y.~Qiao, ``Internimage: Exploring large-scale vision foundation
  models with deformable convolutions,''   Proceedings of the IEEE/CVF
  Conference on Computer Vision and Pattern Recognition (CVPR),
  pp.~14\,408-14\,419,  2023.

\bibitem{ravi2024sam2}
N.~Ravi, V.~Gabeur, Y.-T. Hu, R.~Hu, C.~Ryali, T.~Ma, H.~Khedr, R.~Rädle,
  C.~Rolland, L.~Gustafson, E.~Mintun, J.~Pan, K.~V. Alwala, N.~Carion, C.-Y.
  Wu, R.~Girshick, P.~Dollár, and C.~Feichtenhofer, ``Sam 2: Segment anything
  in images and videos,''   International Conference on Learning
  Representations (ICLR),  2025.

\bibitem{carion2025sam3}
N.~Carion, L.~Gustafson, Y.-T. Hu, S.~Debnath, R.~Hu, P.~Sun et~al., ``Sam 3:
  Segment anything with concepts,''   International Conference on Learning
  Representations (ICLR),  2026.

\bibitem{cheng_masked-attention_2022}
B.~Cheng, I.~Misra, A.~G. Schwing, A.~Kirillov, and R.~Girdhar,
  ``Masked-attention {Mask} {Transformer} for {Universal} {Image}
  {Segmentation},''   Proceedings of the IEEE/CVF Conference on Computer Vision
  and Pattern Recognition (CVPR), pp.~1280-1289,  2022.
  https://doi.org/\url{10.1109/CVPR52688.2022.00135}

\bibitem{jain2023oneformer}
J.~Jain, J.~Li, M.~Chiu, A.~Hassani, N.~Orlov, and H.~Shi, ``Oneformer: One
  transformer to rule universal image segmentation,'' Proceedings of the
  IEEE/CVF Conference on Computer Vision and Pattern Recognition (CVPR),  2023.

\bibitem{Rozumnyi_2021}
D.~Rozumnyi, J.~Matas, F.~Sroubek, M.~Pollefeys, and M.~R. Oswald, ``Fmodetect:
  Robust detection of fast moving objects,''   Proceedings of the IEEE/CVF
  International Conference on Computer Vision (ICCV), pp.~3521-3529,  2021.
  https://doi.org/\url{10.1109/iccv48922.2021.00352}

\bibitem{CHAPEL2020100310}
M.-N. Chapel and T.~Bouwmans, ``Moving objects detection with a moving camera:
  A comprehensive review,'' Computer Science Review, Vol.38, p.~100310,  2020.
  https://doi.org/\url{10.1016/j.cosrev.2020.100310}

\bibitem{Hou2023survey}
B.~Hou, Y.~Liu, N.~Ling, Y.~Ren, and L.~Liu, ``A survey of efficient deep
  learning models for moving object segmentation,'' APSIPA Transactions on
  Signal and Information Processing, Vol.12, No.1, pp.~1-84,  2023.
  https://doi.org/\url{10.1561/116.00000140}

\bibitem{ZHAO202228}
X.~Zhao, G.~Wang, Z.~He, and H.~Jiang, ``A survey of moving object detection
  methods: A practical perspective,'' Neurocomputing, Vol.~503, pp.~28-48,
  2022. https://doi.org/\url{10.1016/j.neucom.2022.06.104}

\bibitem{Gibson1950}
J.~J. Gibson, The Perception of the Visual World.\hskip 1em plus 0.5em minus
  0.4em\relax Houghton Mifflin, 1950.

\bibitem{9413211}
G.~Rahmon, F.~Bunyak, G.~Seetharaman, and K.~Palaniappan, ``Motion u-net:
  Multi-cue encoder-decoder network for motion segmentation,''   Proceedings of
  the 2020 25th International Conference on Pattern Recognition (ICPR),
  pp.~8125-8132,  2021. https://doi.org/\url{10.1109/ICPR48806.2021.9413211}

\bibitem{seong2021hierarchical}
H.~Seong, S.~W. Oh, J.-Y. Lee, S.~Lee, S.~Lee, and E.~Kim, ``Hierarchical
  memory matching network for video object segmentation,''   Proceedings of the
  IEEE/CVF International Conference on Computer Vision (ICCV),  2021.

\bibitem{Yang2021AOT}
Z.~Yang, Y.~Wei, and Y.~Yang, ``Associating objects with transformers for video
  object segmentation,''   Advances in Neural Information Processing Systems
  (NeurIPS), Vol.34, pp.~2491-2502,  2021.

\bibitem{Oh2019STM}
S.~W. Oh, J.-Y. Lee, N.~Xu, and S.~J. Kim, ``Video object segmentation using
  space-time memory networks,''   Proceedings of the IEEE/CVF International
  Conference on Computer Vision (ICCV), pp.~9225-9234,  2019.
  https://doi.org/\url{10.1109/ICCV.2019.00932}

\bibitem{Homeyer2023M3Former}
C.~Homeyer and C.~Schn\"{o}rr, ``On moving object segmentation from monocular
  video with transformers,''   Proceedings of the IEEE/CVF International
  Conference on Computer Vision (ICCV) Workshops, pp.~880-891,  2023.
  https://doi.org/\url{10.1109/ICCVW60793.2023.00098}

\bibitem{Teed2021RAFT3D}
Z.~Teed and J.~Deng, ``{RAFT-3D}: Scene flow using rigid-motion embeddings,''
  Proceedings of the IEEE/CVF Conference on Computer Vision and Pattern
  Recognition (CVPR), pp.~8375-8384,  2021.

\bibitem{Ranftl2021DPT}
R.~Ranftl, A.~Bochkovskiy, and V.~Koltun, ``Vision transformers for dense
  prediction,''   Proceedings of the IEEE/CVF International Conference on
  Computer Vision (ICCV), pp.~12\,179-12\,188,  2021.

\bibitem{He2016ResNet}
K.~He, X.~Zhang, S.~Ren, and J.~Sun, ``Deep residual learning for image
  recognition,''   Proceedings of the IEEE Conference on Computer Vision and
  Pattern Recognition (CVPR), pp.~770-778,  2016.

\bibitem{izumiya2002moving}
K.~Izumida, K.~Shiiya, H.~Takahashi, and S.~Derrouich, ``Moving objects
  detection from travelling monocular camera image,'' IEEJ Transactions on
  Electronics, Information and Systems, Vol.~122, No.3, pp.~498-505,  2002.
  https://doi.org/\url{10.1541/ieejeiss1987.122.3_498}

\bibitem{NEGAHDARIPOUR1989303}
S.~Negahdaripour and B.~K. Horn, ``A direct method for locating the focus of
  expansion,'' Computer Vision, Graphics, and Image Processing, Vol.46, No.3,
  pp.~303-326,  1989. https://doi.org/\url{10.1016/0734-189X(89)90035-2}

\bibitem{Perazzi2016}
F.~Perazzi, J.~Pont-Tuset, B.~McWilliams, L.~{Van Gool}, M.~Gross, and
  A.~Sorkine-Hornung, ``A benchmark dataset and evaluation methodology for
  video object segmentation,''   Proceedings of the IEEE/CVF Conference on
  Computer Vision and Pattern Recognition (CVPR),  2016.

\bibitem{OB14b}
P.~Ochs, J.~Malik, and T.~Brox, ``Segmentation of moving objects by long term
  video analysis,'' IEEE Transactions on Pattern Analysis and Machine
  Intelligence, Vol.36, No.6, pp.~1187 - 1200,  2014, preprint.

\end{thebibliography}
\end{document}